# Advancing Object Detection in Transportation with Multimodal Large Language Models (MLLMs): A Comprehensive Review and Empirical Testing

Huthaifa I. Ashqar
Civil Engineering Department
Arab American University
13 Zababdeh, Jenin P.O Box 240, Palestine
Artificial Intelligence Program
Fu Foundation School of Engineering and Applied Science
Columbia University
500 W 120th St, New York, NY 10027, United States
huthaifa.ashqar@aaup.edu

Ahmed Jaber*
Department of Transport Technology and Economics
Budapest University of Technology and Economics
Műegyetem rkp. 3., H-1111 Budapest, Hungary
ahjaber6@edu.bme.hu

Taqwa I. Alhadidi
Civil Engineering Department
Al-Ahliyya Amman University
Al-Saro Al-Salt, Amman 19111, Jordan
t.alhadidi@ammanu.edu.jo

Mohammed Elhenawy
CARRS-Q
Queensland University of Technology
130 Victoria Park Rd, Kelvin Grove QLD 4059, Australia
mohammed.elhenawy@qut.edu.au

*Corresponding Author

# Advancing Object Detection in Transportation with Multimodal Large Language Models (MLLMs): A Comprehensive Review and Empirical Testing

## Abstract

This study aims to comprehensively review and empirically evaluate the application of multimodal large language models (MLLMs) and Large Vision Models (VLMs) in object detection for transportation systems. In the first fold, we provide a background about the potential benefits of MLLMs in transportation applications and conduct a comprehensive review of current MLLM technologies in previous studies. We highlight their effectiveness and limitations in object detection within various transportation scenarios. The second fold involves providing an overview of the taxonomy of end-to-end object detection in transportation applications and future directions. Building on this, we proposed empirical analysis for testing MLLMs on three real-world transportation problems that include object detection tasks namely, road safety attributes extraction, safety-critical event detection, and visual reasoning of thermal images. Our findings provide a detailed assessment of MLLM performance, uncovering both strengths and areas for improvement. Finally, we discuss practical limitations and challenges of MLLMs in enhancing object detection in transportation, thereby offering a roadmap for future research and development in this critical area.

## Introduction

Object detection is becoming more significant in the field of transportation because it is of critical importance in terms of transportation security, efficiency, and safety. Object detection is the methodology of which is used to identify the items and categorize them, including vehicles, traffic signs, pedestrians, etc., which are needed for several applications including autonomous driving, traveler behavior analysis, and urban infrastructure planning [1], [2], [3]. In autonomous driving, reliable object detection is the most important method to navigate safely and avoid crashes and traffic signal violations. In traffic monitoring, it provides on-the-spot traffic flow analysis, dealing with congestion, and real-time incident detection. Accurate object detection also allows urban planners and traffic controllers to build better infrastructures by understanding the different ways people and traffic move in an area. Transportation is a sector with increasing activity as urbanization and population growth continue, and the accurate detection and interpretation of objects within these systems becomes significantly essential [4]. The results of enhanced object detection capabilities can lead to significant reductions in crash rates, optimized traffic flow, and better urban planning, thereby substantially contributing to the overall improvement of transportation systems.

Multimodal Large Language Models (MLLMs) represent a breakthrough advancement as an artificial intelligent (AI) model that has the ability to integrate multimodal data processing capabilities [5]. Specifically, the models that use their vision capability is called in the literature Large Vision Models (VLMs), which uses texts and images as input with relatively less reasoning capabilities. MLLMs can handle and analyze data from several sources including text, images, videos, and sensor data to provide a comprehensive understanding of complex environments in different settings [6], [7], [8], [9]. MLLMs are built with advanced architectures including transformers, which concurrently process multiple data input streams along with their interpretation. It might lead them to the point of achieving tasks that require a combination of linguistic and perceptual knowledge. The combination permits MLLMs to understand and deal with challenging situations as effectively as real-life single-model scenarios [10]. As shown in Figure 1, a general architecture of an MLLM includes three components. Modality encoder, which condenses raw data into a more streamlined representation that can be used by the AI model. This process utilizes a pre-trained encoder (e.g., CLIP) that has been calibrated to other modalities. Large Language Model (LLM) backbone is the second component, which is required to output responses in text. The third component is the modality interface, which can be used as a link between the encoder and the LLM components as LLMs can only interpret text.

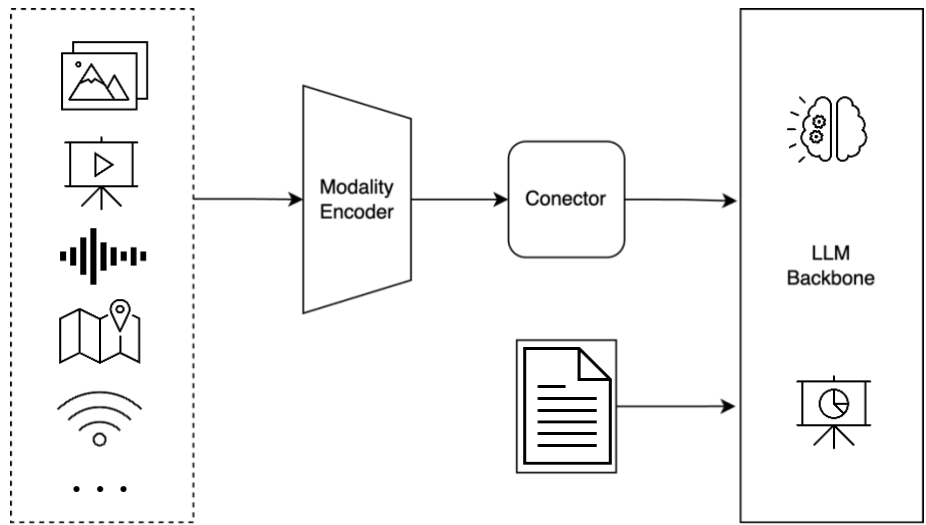

*Figure 1. A general architecture of an MLLM.*

In transportation, MLLMs has the potential to combine visual data from cameras with textual data from traffic reports and sensor data from vehicles to create a more accurate and comprehensive understanding of the environment as well as provide real-time recommendations as a user-friendly output [11]. This ability to integrate and interpret multiple data types and sources can uniquely position MLLMs to handle the complexities of object detection for advanced transportation systems, where diverse data streams are constantly generated and need to be analyzed in real-time.

## Would MLLM Be Leading the Future of Object Detection?

MLLMs can potentially offer several advantages over traditional object detection models. One of the primary advantages of using MLLMs is their flexibility in understanding various complex settings and environments without the need for bounding boxes, which can be a hurdle for traditional object detection models [5]. MLLMs also effectively handle diverse data types, resources, and tasks, Which make them adaptable to various transportation scenarios and tasks [12]. Moreover, MLLMs have wider capabilities as they can search, locate data simply, and broad their understanding from a different type of information and especially in visual form [13], [14], [15]. Utilizing multimodal data, MLLMs are able to have a higher accuracy rate for object detection tasks versus the models that are based on a particular data type. This potential of improved accuracy is so crucially important for applications such as autonomous driving where the detection of objects is necessary for safe driving [13], [14], [15]. MLLMs also offer enhanced contextual understanding as they can interpret complex scenes by combining information from multiple sources, providing a more profound contextual understanding. This capability enables them to make more informed decisions and recommendations such as recognizing a pedestrian in a crosswalk even if partially obscured by other objects [16], [17], [18].

Scalability and transferability can be regarded as another advantageous point. MLLMs can be brought to the scale of large data sets processing and applications working in real-time, and that is crucial in dynamic transportation facilities. These data can be transformed into AI-based system solutions. The support of scalability is, as a result, they can endure the enormous quantity of data in modern transportation systems [19], [20]. Another positive aspect of language models is that they decrease the need for huge amounts of training data, which is one of the main problems with machine learning systems. MLLMs, thanks to abilities such as zero-shot and few-shot learnings, can work outstandingly even with insufficient training data, which in turn diminishes the dependence on big and annotated datasets. This can make MLLMs a potential cost and time-saving alternative in the context of object detection in transportation [21], [22], [23].

MLLMs may provide more accurate and reliable object detection capabilities for transportation applications. It becomes possible to use them in location estimation, high-resolution image matching, and object recognition algorithms that increase the transportation system performance by faster and more effective data processing and collaboration [24]. In autonomous driving, MLLMs could enhance the perception system of the vehicle for real-time and correct detection and reaction to road signs, vehicles, and pedestrians [10]. Better detection capability is paramount for instant decision-making that ensures safety and efficiency in vehicle operation. MLLMs will be able to integrate data from cameras, sensors, and reports in monitoring traffic to obtain detail about the current state of the traffic, hence better management and incident responses in the future [25], [26]. Contextual understanding is superior; therefore, it recognizes patterns and anomalies in traffic flow, hence improving safety and efficiency. Moreover, MLLMs can guide city planners with elaborate insights into traffic patterns and pedestrian behaviors, thereby giving planners better infrastructure and optimized strategies related to traffic management [27], [28]. By leveraging the strengths of MLLMs, transportation systems can achieve higher levels of performance and reliability, addressing the growing demands and complexities of modern urban environments. Besides, connectivity between MLLMs and transportation programs offers a promising future to the technologies for ensuring safer and more convenient transportation.

## Study Contribution

In this paper, we present a comprehensive review study that discusses the use of MLLMs for object detection applications in transportation systems. As opposed to the traditional approaches directed to single-modality data, our study is about the integration of multimodal data streams to improve object detection capabilities. Our work aims to highlight the transformative potential of MLLMs in transportation systems with the help of an extensive review of current technologies but also presenting empirical tests. The intensity, effectiveness, and safety of transportation systems can be improved by MLLMs more effectively when the accuracy is high, hence we analyze the theoretical and practical aspects of this topic. We stimulate new research and innovation in the area and promote deeper interaction of the MLLMs with the end-to-end object detection paradigm through this study. The main contribution of this study includes:

1. Provides a detailed review of MLLMs and VLMs in transportation object detection, highlighting benefits and limitations.
2. Introduces a structured taxonomy for end-to-end object detection methods in transportation using MLLMs.
3. Proposes future directions and applications for MLLMs in transportation object detection.

4. Conducts real-world empirical tests on MLLMs for three transportation problems: road safety attribute extraction, safety-critical event detection, and visual reasoning of thermal images.
5. Highlights the potential of MLLMs to enhance intelligent transportation systems, contributing to safer, more efficient AVs and traffic management.
6. Identifies key challenges and limitations in MLLMs such as order compositional understanding, fine-detail recognition, object hallucination, and computational limitations.

## Existing Object Detection Technologies

Traditional methods for object detection in the transportation sector have historically relied on manual bounding box annotations in input images [29], followed by the extraction of features such as gradient histogram features, scale invariant features, and Haar-like features [30]. These features are then classified to determine the class of the object being detected. Approaches like the YOLO series algorithms and SSDs have introduced regression-based object detection algorithms that are more efficient and accurate, such as the YOLO-MFE method that utilizes multiscale feature extraction for improved accuracy [30]. These newer algorithms leverage deep learning techniques and neural networks to enhance object detection performance. For example, object detection using SSD and MobileNets has been shown to be efficient in quickly detecting objects with fewer resources while maintaining high performance [31]. The development of advanced object detection algorithms has led to the creation of specialized methods that cater to specific applications in transportation, such as the RRPN algorithm designed for object detection in autonomous driving vehicles using radar-based real-time region proposal networks [32]. These specialized algorithms showcase the progression of object detection techniques to fulfill unique requirements of transportation systems, particularly in safety-critical applications, such as autonomous driving.

Several traditional techniques were used to detect objects including CNN, Yolo, and vision transformers. CNN was used extensively in detecting different objects using images and videos. Researchers enhance the CNN performance by 44.6% when they used maximal clique algorithm. The improved CNN shows its capability of detecting small objects [33]. In other works, researchers proposed to use signals with different lengths with the lighter Convolutional neural networks, without using the heavier Recurrent Neural Networks (RNN), which achieved a comparable performance. Results indicated that the modified CNN reduced the number parameters and improves the processing methods [34]. Another enhancement on CNN was done by incorporating the classic CNN and hand-crafted features extraction was done to assist drivers including HOG, ICF and ACF. The results of the work improved object detection for ADAS [35].

Yolo was used extensively in detecting objects including traffic signs [36], pavement cracks [37], traffic scenes [38], and unmanned vehicles [39]. Yolov4 was enhanced by using SwinT to detect cars and persons in different traffic scenes. Model performance showed a high prediction accuracy where the cars detection precision was 89% and the persons detection precision was 94%. In using the real-time application, the YOLOv5 was modified and used to detect ship in real-time. The modified YOLOv5 showed an improvement of the traditional YOLOv5 by 1.3%. Model average precision was 96.6% under different detection schemes [39]. Researchers proposed a YOLO-MXANet for small object detection in traffic scenes, offering improved accuracy, reduced complexity, and faster detection speed compared to traditional methods. Ding et. al [40] proposed an anti-disturbance and variable-scale spatial context features (AVD) detector for road detection, where the training of the multi-layer features of the detector is always taken under the imposing of fake-feature-disturbance from an independent generator, which is trained to exacerbate the detector errors and the mistakes of feature discriminator. The results indicated that the proposed method improves road detection accuracy by 3% on the Munich remote sensing dataset and 0.4% on the urban road dataset [40].

A novel deep learning anchor-free approach based on CenterNet using the Atrous Spatial Pyramid Pooling (ASPP) to improve the detection performance while not increasing the computational cost and the number of parameters. A large-scale naturalistic driving dataset (BDD100K) was used to examine the effectiveness of the proposed approach. The experimental results show that the proposed approach can effectively improve the detection performance on small objects in various traffic situations [41].

### Limitations of Existing Object Detection Technologies

However, these traditional methods have limitations in terms of efficiency and accuracy. For instance, in complex railway scenes, traditional object detection approaches may be inefficient or lack the necessary accuracy, especially when dealing with small objects [42]. Additionally, traditional object detection algorithms often use horizontal bounding boxes to label objects in images, which can lead to accuracy issues and include excessive background information[43]. Moreover, some traditional object detection systems rely on static traffic cameras or expensive mobile units for deployment, which can be costly to establish and maintain or lack diversity in deployment options [44]. These systems may not be as adaptable or cost-effective as desired for widespread transportation applications. Furthermore, traditional methods may struggle with detecting objects with arbitrary orientations, as they typically use horizontal boxes for object labeling, potentially leading to accuracy loss and increased background interference [43].

## Object Detection Using MLLMs

Recently, the integration of MLLMs and VLMs into object detection systems has seen growing interest, especially in the transportation engineering field. The ability of MLLMs to process diverse types of data, such as images, videos, and text, opens new avenues for improving the accuracy and context-awareness of object detection in critical domains like autonomous driving and traffic safety. We provide in the following comprehensive literature review of the research that used MLLMs, VLMs, and LLMs for object detection in transportation engineering. The methodology for this comprehensive review is illustrated in Figure 2.

In the Identification phase, we began by conducting a comprehensive search using Google Scholar. We used specific search terms such as "MLLM or LLM," "VLM," "object detection," and "transportation engineering" to ensure we captured a wide range of relevant studies. This initial search yielded around 782 articles, encompassing various approaches and methodologies in object detection within transportation systems or other fields.

During the screening phase, we aimed to refine the list of articles to focus on the most relevant and high-quality studies. First, we excluded preprints to concentrate on peer-reviewed and published work. Following this, we conducted a title screening to eliminate articles that did not directly relate to our topic, which reduced the pool to about 116 articles. Next, we performed an abstract screening, which involved a more detailed evaluation of each article's abstract to ensure they aligned with our specific interest in MLLM and object detection. This step further narrowed the list to about 25 articles that were highly relevant to our research.

Finally, in the inclusion phase, we carefully evaluated the full text of the eligible articles. This step involved a thorough review of each paper to determine whether it specifically employed MLLM in the context of object detection within the transportation engineering field. After this detailed evaluation, we identified about 8 key articles that directly applied MLLM for object detection, making them highly relevant to our research objectives. These final articles will form the basis of our analysis and review.

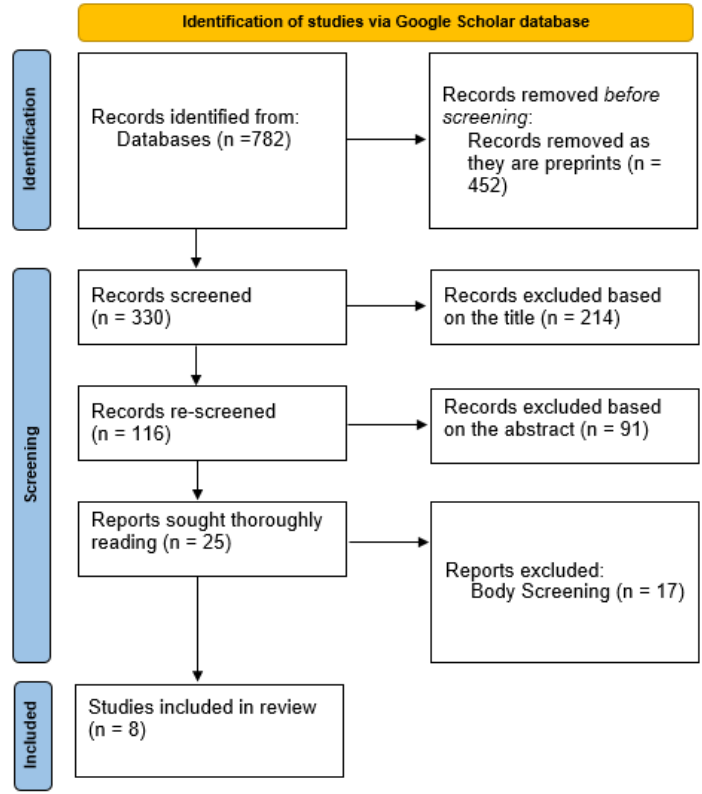

Figure 2. Methodology of literature review in this study.

## Focused State-of-Art

### *Relevance to Transportation*

In the following, we describe the common aspects and differences to thoroughly discuss the use of MLLMs in object detection for transportation engineering. Table 1 shows these eight articles, their taxonomy, and relevance to transportation engineering.

Table 1. The relevance of the focused state-of-art to transportation applications.

| Study | Title | Taxonomy | Transportation Relevance |
|---|---|---|---|
| [45] | Multimodal Driver Condition Monitoring System Operating in the Far-Infrared Spectrum | Driver fatigue detection via thermal images | High relevance for transportation safety |
| [6] | Using Multimodal LLMs for Automated Detection of Traffic Safety-Critical Events | Safety-critical traffic events detection | High relevance for road safety and potential hazards |
| [46] | GPT-4 Enhanced Multimodal Grounding for Autonomous Driving: Leveraging Cross-Modal | Object detection for autonomous driving | High relevance for autonomous driving systems |

| | Attention with Large Language Models | | |
|---|---|---|---|
| [47] | Semantic Scene Understanding with Large Language Models on Unmanned Aerial Vehicles | Scene understanding for Unmanned Aerial Vehicles | Indirect relevance but a potential application in aerial traffic monitoring |
| [48] | Contextual Object Detection with Multimodal Large Language Models | General object detection in human-AI interaction scenarios | Indirect relevance but a potential for object detection in varied contexts |
| [49] | Multi-Modal GPT-4 Aided Action Planning and Reasoning for Self-driving Vehicles | Explainable decision-making from monocular cameras | High relevance for planning and reasoning in self-driving vehicles |
| [50] | OpenAnnotate2: Multi-Modal Auto-Annotating for Autonomous Driving | Integrates various modalities to improve annotation accuracy for large datasets | High relevance for auto-annotating data in autonomous driving systems |
| [51] | Pix2Planning: End-to-End Planning by Vision-language Model for Autonomous Driving on Carla Simulator | End-to-end vision-language modeling for autonomous driving | High relevance for translation visual inputs into language-based trajectory planning |

As shown in Table 1, Knapik et al. [45] and Tami et al. [6] emphasizes transportation safety, with Knapik focusing on driver fatigue detection using thermal imaging and CNNs, while Tami et al. addresses automated detection of traffic safety-critical events. Both are highly relevant to improving road and driver safety. Liao et al. [46] explores object detection for autonomous driving using GPT-4 and Vision Transformer models, enhancing the safety and efficiency of autonomous vehicles. However, de Curtò et al. [47] investigates scene understanding using MLLMs on UAVs, which could have potential applications in aerial traffic monitoring, though its transportation relevance is indirect. Similarly, Zang et al. [48] focuses on general object detection in human-AI interaction scenarios, with indirect transportation relevance in broader object detection contexts. Additionally, Chi et al. [49] investigates the use of GPT-4 for action planning and reasoning in self-driving vehicles. Authors focused on explainable decision-making using multi-modal data from monocular cameras and demonstrated how LLMs can provide human-understandable driving rationales. Continually, Zhou et al. [50] presents OpenAnnotate2, a multi-modal tool for auto-annotating data in autonomous driving systems, while Mu et al. [51] proposes Pix2Planning, an end-to-end vision-language model for autonomous driving, which translates visual inputs into language-based trajectory planning. The system uses bird's-eye view (BEV) features to enhance planning precision on the CARLA simulator.

### *Model and Architectures*

As Table 2 shows, de Curtò et al. [47], the study used YOLOv7 with CLIP prefix captioning for object detection from UAVs, while Knapik et al. [45] incorporated far infrared and visual inputs using various sensors and multimodal fusion. Tami et al. [6] deployed a framework that integrates textual, visual, and auditory modalities to detect safety-critical events using models like Gemini-Pro Vision 1.5. However, Zang et al. [48] highlighted zero-shot detection models for road detection using a pipeline of CLIP and object detectors. Finally, Liao et al. [46] focused on autonomous driving by leveraging GPT-4's capabilities for grounding vision and language, as well as Chi et al. [49] and Zhou et al. [50].

It was found that the most common architectures are GPT-based models and CNN such as YOLOv8 as well as Vision Transformers, which provide high accuracy in object detection. It was noted that fine-tuning was essential for models that required real-time performance, particularly in the Multimodal Driver Condition Monitoring System and the Traffic Safety-Critical Event Detection.

### *Datasets Description*

As Table 2 shows, the datasets used in the analyzed studies vary widely in size and data type. Public datasets such as KITTI and COCO are frequently used, particularly for autonomous vehicle perception tasks. Custom datasets are often used for specific applications like driver monitoring and UAV-based scene understanding.

*Table 2. Overview of the models and datasets used in previous studies.*

| Study | Dataset | Data Type | MLLM/LLM Used | Size | Performance |
|---|---|---|---|---|---|
| [45] | Thermal image dataset | Images | ChatGPT3.5-turbo + YOLOv8 | Large | mAP = 0.94 |
| [6] | DRAMA dataset | Images, Text | Gemini-Pro-Vision 1.5 MLLM | Medium | Accuracy = 79% |
| [46] | Talk2Car | Images, Text | GPT-4, Vision Transformer | Large | mAP = 0.75 |
| [47] | Custom UAV dataset | Images, Video | GPT-3, CLIP, YOLOv7 | Medium | - |
| [48] | COCO, CODE Benchmark | Images, Text | GPT-3, BLIP-2 | Large | AP = 43.4 |
| [49] | CARLA | Images, Text, and sensor data | GPT-4, Graph-of-Thought (GoT) | Large | - |
| [50] | SemanticKITTI | Images and multi-modal sensor data | GPT-4, Langchain | Large | - |
| [51] | CARLA | images, BEV representations, and text | Auto-regressive transformer | Large | DS = 92.39 RC = 99.60 |

### *Performance Metrices*

The studies reported various performance metrics, with precision being the most commonly reported. As Table 2 shows, using MLLMs for traffic safety-critical events achieved a high accuracy of about 79% in detecting safety-critical events using few-shot learning, outperforming older models [6]. Semantic scene understanding with UAVs demonstrated success in zero-shot scene descriptions but faces limitations in handling complex and crowded scenes. Contextual object detection highlighted how combining multimodal inputs can improve detection accuracy, though model performance depends on data quality. The GPT-4 enhanced multimodal grounding study achieved its highest mAP (mean Average Precision) at 75%. Furthermore, Chi et al. [49] exhibited a strong performance in generating accurate, interpretable actions with high decision-making transparency, while Zhou et al. [50] showed notable improvements in data annotation efficiency and accuracy for autonomous driving datasets, but detailed performance metrics such as precision or recall were not reported. Finally, Pix2Planning achieved a strong performance on CARLA benchmarks, with a Driving Score (DS) of 92.39 and Route Completion (RC) of 99.60, outperforming other visual-only methods by leveraging the auto-regressive transformer to enhance planning precision.

*Model Complexity*

The model complexity across the eight reviewed articles varies depending on the tasks and modalities integrated into the detection systems. In Knapik et al., the complexity stems from combining thermal imaging and visual data using CNNs and MLLMs, which adds to the computational load due to real-time processing and multimodal fusion of far-infrared and visual inputs. Tami et al. presented a framework that integrates textual, visual, and audio data for detecting traffic safety-critical events [6]. The use of object-level question-answer prompts in the MLLM increases complexity by requiring the model to synthesize data from multiple sources for accurate hazard detection [52], [53], [54]. Liao et al. incorporated GPT-4 and Vision Transformer models for object detection in autonomous driving, resulting in high model complexity due to the cross-modal attention mechanisms and large-scale pre-training required to understand complex driving environments. Moreover, although de Curtò et al. focused on UAVs, their study involved significant complexity in scene understanding using MLLMs, which require zero-shot learning and real-time data processing. However, Zang et al. explored contextual object detection in varied scenarios, further increasing complexity by integrating multimodal data for human-AI interaction, requiring the models to process diverse data inputs and provide contextualized responses. Chi et al. increased model complexity through its use of the Graph-of-Thought (GoT) structure, adding substantial computational overhead due to its interpretability and explainability mechanisms. In contract, Zhou et al. simplified the annotation process using a deep neural network for multi-modal auto-annotating, though it introduces some complexity due to handling large-scale multi-sensor data. Mu et al. added significant computational complexity due to the large receptive field and the transformer's ability to maintain spatial information over long sequences. Nonetheless, the integration of multiple data modalities and the need for real-time, high-accuracy detection contributes to the high computational and architectural complexity across these studies.

*Prompt Engineering*

Prompt engineering is pivotal across the eight studies for enhancing MLLM performance. Knapik et al. used tailored prompts to interpret far-infrared images and detect driver fatigue, while Tami et al. employed object-level question-answer prompts to focus on identifying traffic safety-critical events [6]. Liao et al., on the other hand, leveraged advanced prompts with GPT-4, guiding the model to detect key elements in autonomous driving scenes. de Curtò et al. used prompts for zero-shot scene descriptions in UAVs, and Zang et al. applied prompts to help MLLMs detect objects in varied human-AI interaction contexts. Finally, Chi et al. used Graph-of-Thought (GoT) prompts while Mu et al. employed a vision-language prompt to translate waypoints into language sequences. These engineered prompts ensured more accurate and relevant object detection across the studies.

*Limitations of MLLMs from Previous Studies*

The following is a list of the limitations reported by the eight studies.

- Knapik et al.: The main limitation lies in the reliance on far-infrared imaging, which may not perform well in diverse weather or lighting conditions. Additionally, the model's ability to generalize to different types of drivers or vehicles is not fully explored as it is Limited to specific driver conditions (e.g., yawning).
- Tami et al.: While the framework effectively detects traffic safety-critical events, its reliance on multimodal data inputs may result in computational inefficiencies, making real-time deployment in complex urban settings a challenge [6].
- Liao et al.: The complexity of using GPT-4 and Vision Transformers for object detection creates significant computational demand, limiting its scalability for real-world autonomous

- driving applications. Furthermore, fine-tuning the model for diverse driving conditions and weather scenarios remains an unresolved challenge.
- de Curtò et al.: The study's focus on UAV scene understanding presents limitations when applied to crowded or fast-changing environments. The zero-shot learning approach may also struggle with novel or highly complex scenes that it has not been pre-trained on.
- Zang et al.: The general focus on human-AI interaction object detection is broad, leading to limitations in transportation-specific contexts. The model's ability to provide contextually relevant outputs in real-time may falter when faced with highly dynamic or ambiguous scenes.
- Chi et al.: The limitations are noted in real-world applicability due to safety concerns in dynamic environments.
- Zhou et al.: The challenges in scaling the tool for larger datasets, especially when dealing with high-dimensional sensor data.
- Mu et al.: The high computational demand of the Pix2Planning model, which may limit real-time deployment.

All papers discussed the challenge of generalizing models across different environments, such as urban vs. rural, weather conditions, or lighting. Some models struggled with small object detection or suffer from hallucinations in text outputs. Additionally, computational complexity and delays in processing multimodal data in real-time are common limitations.

## Not Focused State-of-Art

While this review primarily focuses on the intersection of MLLMs/LLMs, object detection, and transportation engineering, there are several relevant studies that do not directly incorporate MLLMs or LLMs but still contribute significantly to advancements in transportation systems and object detection. Below are summaries of key studies that, while not the primary focus, offer valuable insights into the broader landscape of object detection in transportation and related fields.

The paper Xi et al. [55] focused on using GSENet (Global Semantic Enhancement Network) for roadway feature interpretation in autonomous driving systems. The primary contribution is enhancing semantic segmentation of road features using integrated domain adaptation methods. The model works by improving the interpretation of unstructured road environments, which is crucial for autonomous vehicles operating in complex urban settings. While the paper uses deep learning for road feature detection, it does not involve MLLMs or LLMs and is focused on semantic segmentation rather than object detection. Similarly, Ou et al. [56] introduced Drone-TOOD, a lightweight object detection model designed specifically for UAVs to detect vehicles from aerial imagery. The model is task-aligned, meaning it is optimized for the unique challenges of object detection in UAV images, such as varying altitudes, angles, and scales. Finally, Alaba et al. [57] examined the use of multimodal fusion techniques to enhance 3D object detection in autonomous vehicles. It explored the integration of data from multiple sensors, such as LiDAR, cameras, and radar, to improve the accuracy and robustness of object detection in autonomous driving systems.

## Future Work from Previous Studies

As previous studies have argued, future research should focus on improving the scalability and robustness of MLLMs for real-world transportation applications. For example, it is needed to handle diverse environmental conditions, reduce computational costs and optimize real-time performance. Also, there are potential improvements by incorporating new sensor types (e.g., radar, LiDAR) or refining multimodal fusion techniques to enhance object detection. Further research on streamlining data annotation tools could also help manage the growing demand for large, high-quality datasets. Finally, reducing the computational complexity of transformer-based models used in end-to-end planning, will be crucial for making these models more practical for real-time deployment in large-scale autonomous driving systems.

## Future Directions and Potential Applications

This section provides a comprehensive taxonomy on the use of MLLMs (or VLMs) within the domain of transportation for object detection and classification. Figure 3 organizes the taxonomy for object detection into around three major, basic, and downstream tasks: Perception and Understanding, Navigation and Planning, and Decision-Making and Control. Each of these tasks represents a core area of object detection for transportation applications including autonomous driving and Intelligent Transportation Systems (ITS). Perception and Understanding includes tasks such as Vulnerable Road Users (VRU) detection and visual scene reasoning for the surrounding environment, which are considered essential for the vehicle to accurately and safely perceive and understand the environment. Navigation and Planning includes the tasks that VLMs assist in guiding the vehicle and providing recommendations about its speed and acceleration such as localization and motion planning. In the area of Decision-Making and Control, MLLMs play an important role in providing decisions on both the open-loop and closed-loop levels. This includes tasks that ensure the vehicle can respond appropriately to real-time scenarios by understanding the environment. The taxonomy in Figure 3 depicts a structured view on how MLLMs can by potentially integrated into different aspects of end-to-end object detection. It shows how these models can contribute to the overall functionality and safety of different transportation tasks.

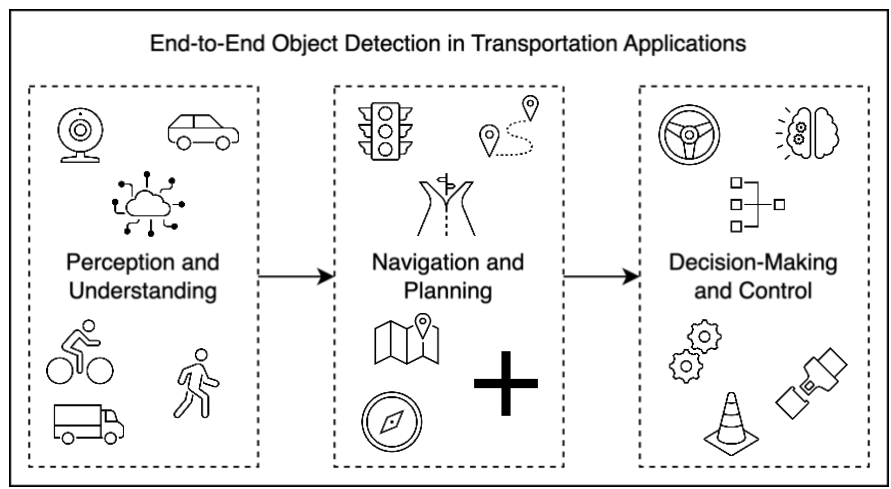

*Figure 3. An overview of the taxonomy of end-to-end object detection in transportation applications.*

To investigate the potential of MLLMs, we present in the next section three different case studies that can be linked to the taxonomy of object detection tasks and applications in autonomous driving and ITS. The first case study, namely Road Safety Attributes Extraction, falls under the Perception and Understanding category and aligns with tasks such as VRU detection and what is called Open-Vocabulary Object Detection (OVD). In OVD, MLLM combines the power of vision and language to provide accurate and actionable descriptions of objects in scenes, including their attributes, behaviors, relations, and interactions. This case study focuses on extracting and understanding various safety-related events from the images, which can be important for perceiving and interpreting the surrounding environment in ITS applications. The second case study, namely Safety-Critical

Event Detection, is closely represents the Decision-Making and Control category. It is related to the areas of open-loop and closed-loop decision making. The detection of safety-critical events requires the MLLM to identify and react to potential hazardous events and provide a user-friendly decision in real-time that ensures the safety and control of the autonomous vehicle and the surrounding environment and traffic. In the third case study, namely, Visual Reasoning of Thermal Images, we investigate the potential of MLLMs within the Navigation and Planning category. In this case study, emphasizes the ability of automated vehicles to navigate during adverse environmental conditions using thermal images. The three cases show MLLMs can contribute to the broader goals of perception, decision-making, and navigation in this field. Our goal is to demonstrate the potential of integrating MLLMs in object detection applications, which highlights some of their advantages in improving accuracy, robustness, and versatility in various scenarios.

## Case Study 1: Road Safety Attributes Extraction

Road safety is a critical concern worldwide, with the International Road Assessment Program (iRAP) playing a pivotal role in evaluating and improving road conditions to reduce accidents and fatalities [58], [59]. The iRAP is an umbrella program for Road Assessment Program (RAPs) worldwide that are working to reduce crashes and save lives using a robust, evidence-based approach to prevent unnecessary deaths and suffering. Nonetheless, traditional assessment methods are labor-intensive and time-consuming [60]. The introduction of Google Street View imagery and advances in MLLMs offer promising alternative [61]. This case study explores the integration of zero-shot in-context learning and MLLMs to automate the extraction of road safety attributes from Google Street View images, providing a scalable and efficient solution for road safety assessments.

In this case study, we collected 168 street view images from various driving environments in Brisbane, Queensland, Australia, encompassing multiple road conditions and features. Utilizing zero-shot in-context learning, we prompted GPT-4-vision-preview to recognize and interpret these images in the context of iRAP safety attributes. This approach allowed the model to generate relevant safety features without explicit prior training on the iRAP dataset, showcasing the potential of MLLMs in understanding complex driving environments and replacing humans in some of the routine inspections. Table 3 shows an example of a utilized prompt for interpreting Google Street View images for eleven selected iRAP safety attributes and the resulted response.

*Table 3. An Example of a Prompt Utilized for Interpreting Google Street View Images to Extract iRAP Safety Attributes.*

> Your role is systematically inspecting an image for various road features.
> ---
> Road Inspection Checklist
> 1. Number of Lanes - Count the number of lanes in the direction closest to the camera. Provide the number without additional comment.
> 2. School Zone Warning - Inspect for school zone warning signs and select the appropriate option:
>     1. School zone flashing beacons
>     2. School zone static signs or road markings
>     3. No school zone warning
>     4. Not applicable (no school at the location)
> 3. School Zone Crossing Supervisor - Observe for the presence of a school zone crossing supervisor and select one:
>     1. Supervisor present at school start and finish times
>     2. Supervisor not present
>     3. Not applicable (no school at the location)

4. Roads that Cars Can Read
- Inspect the following and provide 'yes', 'no', or applicable comment:
    - Clear and consistent road signage visible and unobstructed.
    - Well-defined road markings, maintained with high contrast for visibility.
5. Roadworks - Examine the area for roadworks and choose one:
    1. No road works
    2. Minor road works in progress
    3. Major road works in progress
6. Pedestrian Fencing - Check for pedestrian fencing (guard rails) and select one:
    1. Not present
    2. Present
7. Vehicle Parking - Inspect the area for vehicle parking and select one:
    1. None
    2. Each road side has at least one parked vehicle
    3. At least one vehicle is parked on one side of the road
8. Speed Management / Traffic Calming - Identify any present speed management features such as curb build-outs, speed humps, raised tables, etc.
9. Median Type
    - Examine the image and identify any present median type:
    - Options include Centre line, Wide centre line, Central hatching, Continuous central turning lane, various physical median widths, safety barriers, etc.
10. Carriageway - Inspect the image and select one:
    1. Divided carriageway
    2. Undivided carriageway
11. Quality of curve - Inspect the image to identify any present road Curvature then select one:
    1. Poor
    2. Not applicable
    3. Adequate

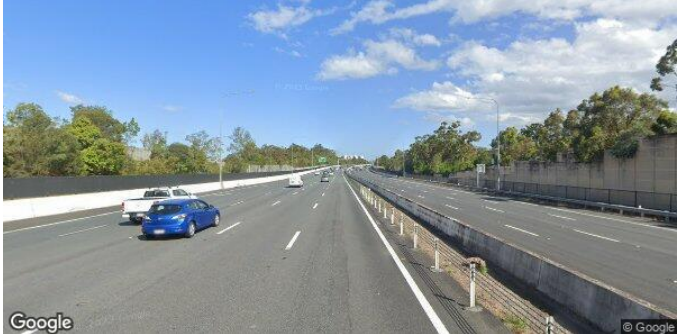

**Number of Lanes:** There are four lanes in the direction closest to the camera.
**School Zone Warning:** No school zone warning
**School Zone Crossing Supervisor:** Not applicable (no school at the location)
**Roads that Cars Can Read:**
Yes, clear and consistent road signage visible and unobstructed.
Yes, well-defined road markings, maintained with high contrast for visibility.
**Roadworks:** No road works
**Pedestrian Fencing:** Present, as there is a barrier between the pedestrian walkway and the road.
**Vehicle Parking:** None
**Speed Management / Traffic Calming: No** visible speed management features such as curb build-outs, speed humps, raised tables, etc.
**Median Type:** Safety barriers are present between the opposite lanes of traffic.

> **Carriageway:** Divided carriageway
> **Quality of Curve:** Not applicable, as no significant road curvature is visible in the image provided.

We then used GPT-4-vision-preview to evaluate the performance in responding to prompts related to eleven selected iRAP criteria for the collected 168 street view images. The model was employed in a zero-shot learning approach, meaning it was not specifically trained on these criteria beforehand. Instead, it was presented with the prompts and required to generate accurate responses based on its base knowledge and understanding. Each response was then assessed for accuracy, with results scaled from 0 to 100. Figure 4 shows the accuracies of a GPT-4-vision-preview model in responding to prompts related to eleven selected iRAP criteria using a zero-shot learning approach.

Results showed that in eight attributes out of eleven, the model was able to achieve an accuracy of more than 80%. However, here are some observations on its performance. The model frequently misidentifies barriers on highways as pedestrian fencing. Emergency bays and shoulders are often mistaken for additional traffic lanes. The model struggles with accurately counting lanes, especially when there are more than three. Moving vehicles on urban streets are sometimes incorrectly classified as parked cars. The model misidentifies the materials of medians. It also confuses low medians for mere road lines.

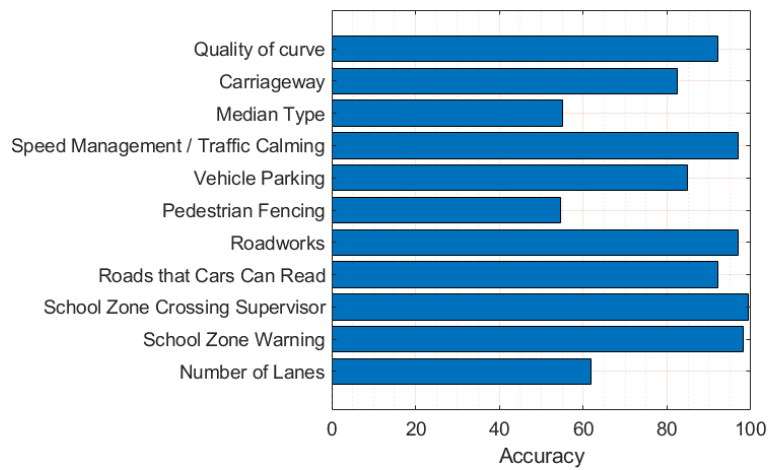

*Figure 4: Analysis of ChatGPT-4's Zero-Shot Learning in Road Feature Recognition.*

This case study introduced a novel approach to extracting critical attributes for the iRAP using street view images combined with zero-shot in-context learning techniques and MLLMs. By analyzing about 168 diverse street view images collected from different driving environments, we demonstrate the promising capability of MLLMs to accurately identify and categorize key road safety features without prior direct training on specific iRAP attributes. This methodology reduces the reliance on extensive labelled datasets. The application of zero-shot learning allows for the flexible adaptation of the model to new attributes or changes in assessment criteria with no additional training. However, the common misclassifications identified signal a need for model enhancements, either through few-shot learning or fine-tuning with a specialized iRAP dataset. Future research will explore few-shot in-context learning, employing a small set of example images to teach the model traffic and road safety concepts more effectively. It needs also evaluate other multimodal models capable of local execution, refining them through supervised learning to tap into the MLLM's inherent foundational knowledge. By incorporating images that delineate the differences between various iRAP attribute values, the goal is to refine the model's accuracy in extracting attributes, thereby setting the stage for more advanced and detailed road safety evaluations.

## Case Study 2: Safety-Critical Event Detection

Recent advancements in MLLMs are enhancing autonomous driving, focusing on textual and multimodal analyses to interpret complex driving environments [62], [63]. These technologies are promising for understanding dynamic road conditions and offering early hazard warnings [64]. Despite their potential, a gap exists in applying these models to safety-critical event analysis. This case study addresses this gap with exploring the use of MLLMs for analyzing safety-critical events in driving, combining text and image analyses to provide illustrative and actionable insights.

We used a multi-stage Question-Answer (Q/A) approach and applied it for two MLLMs including Gemini-pro-vision 1.0 and Llava-7B 1.5. We started with frame extraction, where the system automatically collects video frames from the ego vehicle's camera at unifrom intervals. The frames were then be tested for hazard detection, where the MLLMs assessed the scene for potential danger. Upon detecting a hazard, the framework used a multilateral Q/A strategy to identify the characteristics of the critical event further. For this stage, we used what is called *What, Which,* and *Where* queries to investigate the object-level details. In the "What" phase, the MLLMs classified the objects detected by the camera, while the "Which" phase involved the MLLMs identifying specific features and attributes of the agents involved in the event, and the final "Where" phase tasks the MLLMs to determine the spatial location and distance of the hazardous agents. To evaluate MLLMs, human experts first assessed ground truth data annotating the events. We collected a sample of videos from a YouTube channel, namely Dash Cam Owners Australia, offering diverse scenarios for testing. Comparative analysis measured the model's performance against this data to determine its accuracy in hazard detection and categorization.

Four examples of the prediction of Gemini-pro-vision 1.0 in zero-shot learning are illustrated in Figure 5. To evaluate the effectiveness of MLLMs in detecting safety-critical events, a series of experiments were carried out using Gemini-pro-vision 1.0 and Llava-7B 1.5. For Gemini, we evaluated the model in two folds. In the first experiment, we used two frames as input to the model, and for the second experiment, we used the video directly with no processing. Table 4 presents a comparative analysis of the performance of three different configurations of vision models—Gemini-pro-vision 1.0 with two-frame input, Llava-7B 1.5 with two-frame input, and Gemini-pro-vision 1.0 with video input—across various stages of the Q/A process. The stages include Risk, Scene, What, Which, Where, and Recommended Action, with corresponding accuracy percentages for each configuration. Gemini-pro-vision 1.0 with video input consistently outperforms the other configurations, achieving the highest accuracy in most categories, particularly in the Scene (100%) and Recommended Action (75%) stages. The overall performance metrics indicate that Gemini-pro-vision 1.0 with video input (74.67%) is superior to both its two-frame input counterpart (71.8%) and the Llava-7B 1.5 model (58.6%). This suggests that utilizing video input can significantly enhance the model's ability to accurately process and respond to queries in dynamic and complex visual scenarios.

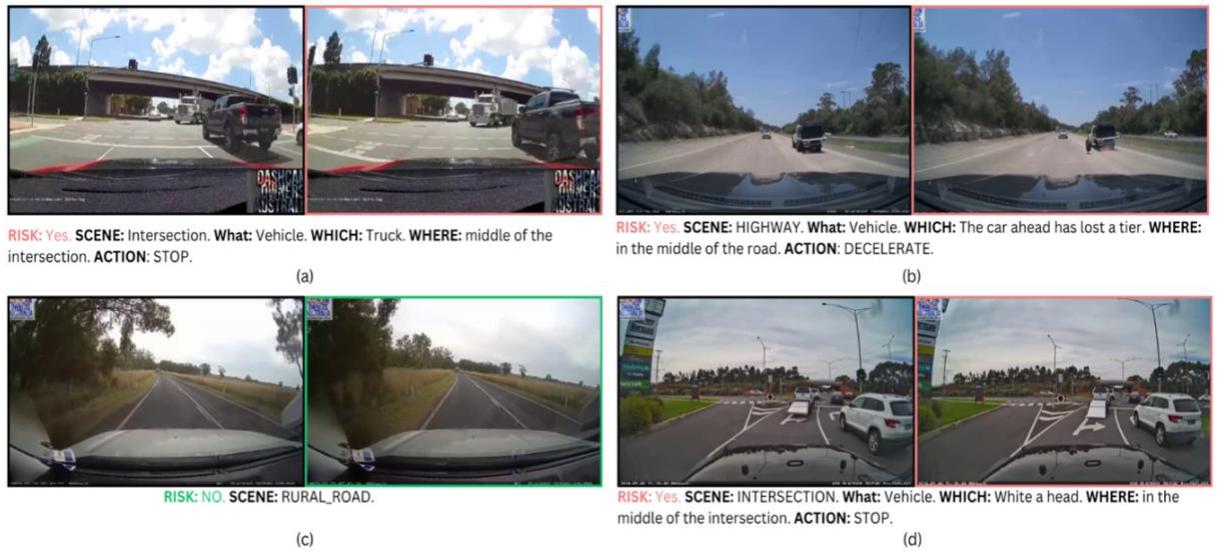

*Figure 5: Four Example Outputs from Gemini-pro-vision Analysis with Two Frames.*

*Table 4: Comparative Performance Analysis of Safety-Critical Event Detection.*

| Q/A Stage | Gemini-pro-vision 1.0 with two-frame input | Llava-7B 1.5 with two-frame input | Gemini-pro-vision 1.0 with video as input |
|---|---|---|---|
| **Risk** | 55% | 46% | 75% |
| **Scene** | 90% | 76% | 100% |
| **What** | 83% | 71% | 66% |
| **Which** | 87% | 59% | 66% |
| **Where** | 61% | 49% | 66% |
| **Recommended Action** | 55% | 51% | 75% |
| **Overall** | **71.8%** | **58.6%** | **74.67%** |

This case study aims to enhance the efficiency of processing naturalistic driving dash cam data, ultimately improving the detection of safety-critical events using an MLLM-empowered model. It demonstrates the potential of leveraging the capabilities of MLLMs in analyzing safety-critical event scenarios using multi-modal data integration and dynamic contextual data reduction for guiding the model's output. Future research in this area will focus on expanding the dataset and scenario range to further validate our innovative approach to driving safety.

## Case Study 3: Visual Reasoning of Thermal Images

Thermal imaging is vital for advancing automatic driving systems due to its ability to detect heat signatures, enhancing the overall perception capabilities of autonomous vehicles [65], [66]. One primary benefit of thermal imaging is its enhanced visibility in adverse conditions in both urban and highway environments [67]. Thermal cameras can detect heat emitted by objects and living beings, providing clear images even in complete darkness, which is essential for detecting pedestrians, animals, and other vehicles at night [68], [69].

This case study explores the capabilities of MLLMs, specifically GPT-4 Vision Preview and Gemini 1.0 Pro Vision, in visual reasoning of thermal images. We evaluate the MLLMs' ability to detect and enumerate objects within thermal images using zero-shot in-context learning. This case study not only shows the potential of MLLMs to enhance object detection accuracy and environmental understanding but also open venues for safer and more reliable autonomous driving systems. We used the Teledyne FLIR Free ADAS Thermal Dataset V2 [70], which includes fully annotated frames covering more than fifteen object classes, captured using a thermal and visible camera pair mounted on a vehicle. The dataset ensures diversity by sampling frames from a wide range of footage, excluding redundant footage, to provide robust training and validation data.

Using zero-shot in-context learning, the models were tested on a sample of the images and demonstrated an ability to process and analyze these modalities. Table 5 shows the performance metrics for GPT-4 Vision Preview and Gemini 1.0 Pro Vision evaluated on four object categories: Person, Bike, Car, and Motorcycle. The metrics include True Positive Rate (TPR) and False Positive Rate (FPR). For person detection, GPT-4 demonstrates a higher TPR (0.57) compared to Gemini (0.39), with both models maintaining an FPR of 0, indicating better accuracy in identifying persons for GPT-4. In bike detection, Gemini shows a slightly higher TPR of about 0.41 than GPT-4 (0.31). For car detection, both models perform well, but Gemini has a slightly higher TPR of 0.90 and a slightly lower FPR of 0.08. In motorcycle detection, Gemini significantly outperforms GPT-4, with a TPR of 0.24 versus 0.08. Overall, the insights suggest that while both models are effective, Gemini 1.0 Pro Vision shows superior performance in bike, car, and motorcycle detection, whereas GPT-4 Vision Preview surpasses in person detection. Detection accuracy also varied among different images and different objects. Although thermal images posed a unique challenge due to their reliance on heat signatures rather than visible light and their reduced visual information, MLLMs achieved moderate success in recognizing objects such as cars and people, establishing a strong foundation for future model development.

*Table 5: Performance Breakdown of GPT4 and Gemini Analyzing Thermal Images.*

| MLLM Model | GPT4 Vision Preview | | Gemini 1.0 Pro Vision | |
|---|---|---|---|---|
| *Evaluation Metrics per Category* | *TPR* | *FPR* | *TPR* | *FPR* |
| Category ID 1: Person | 0.57 | 0 | 0.39 | 0 |
| Category ID 2: Bike | 0.31 | 0.01 | 0.41 | 0.02 |
| Category ID 3: Car | 0.86 | 0.09 | 0.90 | 0.08 |
| Category ID 4: Motorcycle | 0.08 | 0 | 0.24 | 0.01 |

This case study shows the potential of MLLMs, specifically GPT-4 Vision Preview and Gemini 1.0 Pro Vision, in processing and analyzing thermal images for applications in autonomous driving and intelligent transportation systems. The results suggest the need for ongoing enhancements to enhance the accuracy of MLLMs, especially in diverse environmental conditions that autonomous vehicles may encounter. The findings validate the ongoing use and enhancement of MLLMs for image-based processing in autonomous driving.

## Challenges and Limitations

In this section, we offer our understanding of the limitations and challenges of using MLLMs for object detection and classification in transportation applications. We derived these limitations and challenges from implementing the three case studies, studying the state-of-the-art of using MLLMs in object detection, as well as from other studies that we published previously or in progress. Figure 6 depicts a summary of the challenges and limitations.

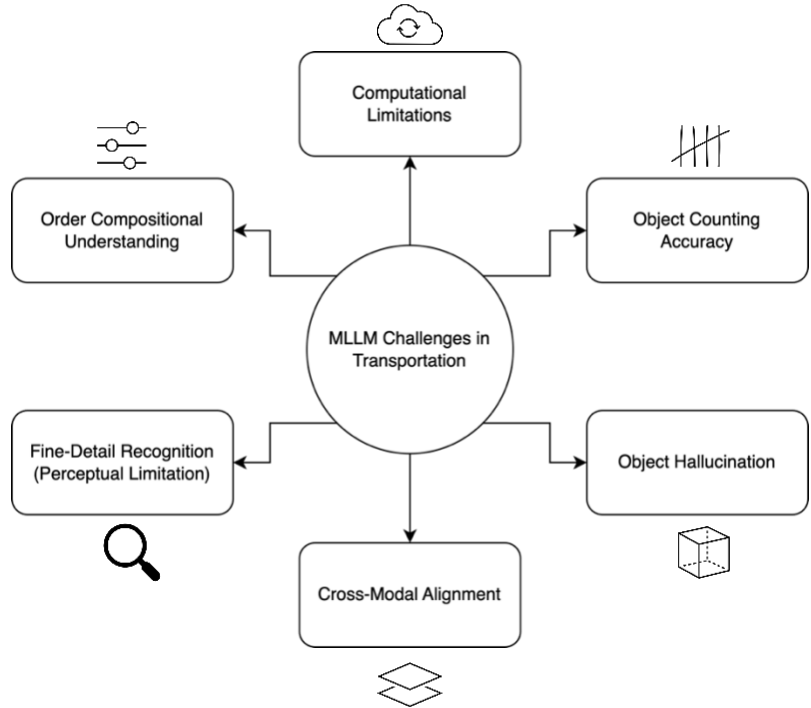

*Figure 6. Challenges and limitations of MLLMs in object detection for transportation applications.*

## Order Compositional Understanding

We found that MLLMs struggle with correctly order and compose the relationships between objects and their attributes in complex scenes [71], [72]. Results showed that these models usually perform well on tasks such as image-text retrieval, but they still frequently show shortages in capturing the precise relationships and the correct order within the visual context. MLLMs seemed to misinterpret and sometimes overlook the importance of the sequence of the scene and in which objects and their corresponding attributes are presented. This frequently led to errors in predicting such as confusing the relationships between objects (e.g., mistaking "the cat on the mat" with "the mat on the cat"). This limitation can be relaxed by more robust training techniques and benchmarking, which ensure that models will accurately interpret and logically reason about the compositional structure inherent in multimodal data.

## Fine-Detail Recognition: Perceptual Limitation

We found that the perceptual limitations of MLLMs were primarily revolved around the struggle of accurately identifying and interpreting small visual objects within images. Although MLLMs showed Promising capabilities in visual question answering (Q/A), their performance (sometime significantly) drops when dealing with relatively smaller objects. This has been majorly influenced by several factors including image and object quality, relative size, its location in the image, and the presence of other surrounding distractors [73]. We also found that even with relatively high-quality images, MLLMs still struggled to recognize smaller objects if they are positioned peripherally or between visual distractors. Additionally, the models' performance varied based on the object's location within the image, which suggested a positional bias in its training.

## Cross-Modal Alignment

Another limitation that we found in the vision encoders of MLLMs was primarily centered upon achieving effective cross-modal alignment and visual correspondence. These two factors were found to be important for the model's performance as they significantly influence how well the vision representations integrate with the language model [74]. Vision encoders might struggle to align visual features with corresponding textual elements and other related modes of data, which made them fail to maintain high fidelity in visual correspondence. This was found particularly when fine-grained details were involved.

## Object Hallucination

In the exploration of MLLM, we found that one notable limitation was object hallucination, in which the model generates objects that do not exist in the target image [75]. This issue presented a significant challenge as it undermined the reliability of MLLMs in some of the tasks and it might become an issue in other applications as well. For example, we found that in scenarios that require high precision and reliability such as Navigation and Planning in AVs, MLLM tends to hallucinate with regard objects, which might lead to unsafe decisions and increase risks. The problem of object hallucination not only highlights a fundamental flaw in the current capabilities of MLLMs but also restricts their practical deployment across sensitive and downstream tasks.

## Object Counting Accuracy

One crucial application for ITS is the counting of objects in the scenes to be used for different applications such as a simple traffic management system. Despite MLLMs' advanced capabilities, we found that MLLMs showed a significant challenge in object enumeration, as shown directly in Case Study 3. The study highlighted how these models can be proficient in identifying objects across diverse modalities such as RGB and thermal imaging but still struggled with precise object counting. This was particularly shown when interpreting complex or overlapping object scenes.

## Computational Limitations

We found that the computational demands for training, fine-tuning, and inference of MLLMs can be substantially driven by the complexity and the scale of these models [76], [77]. We found that zero-shot learning, which does not require training and depends on the knowledge base of the MLLM, showed relatively low performance. We also found that few-shot learning, training, and fine-tuning showed a higher performance. Nonetheless, training an MLLM often involves managing vast amounts of data across multiple modalities (e.g., text, images, and video). This requires extensive computational power involving GPUs or TPUs over long periods. For instance, we found that in some cases the pre-training phase alone can consume hundreds of thousands of GPU hours, depending on the model size and dataset. Fine-tuning these models also demands significant resources, though it is generally more efficient than training from scratch. The fine-tuning process can be optimized through techniques such as quantization and distributed training, which help to reduce memory usage and computational load [78]. Moreover, we found that inference also presents challenges, especially when deploying these models in real-time applications. The memory requirements for managing key-value caches during inference are a major bottleneck [79]. Techniques such as PagedAttention, which optimize memory allocation by using non-contiguous memory blocks, can help alleviate some of these constraints.

# Conclusion

This study provides a comprehensive evaluation of MLLMs and VLMs in the context of object detection for transportation systems. Through an extensive review of current MLLM technologies, we highlight their significant potential in enhancing transportation applications, particularly in tasks such as road safety monitoring and traffic management. While MLLMs and VLMs hold considerable potential for advancing autonomous driving and ITS, there remains a gap between their current capabilities and the practical demands of real-time, safety-critical applications.

The empirical analysis across three real-world transportation scenarios—road safety attributes extraction, safety-critical event detection, and visual reasoning of thermal images—offers valuable insights into the strengths and limitations of MLLMs. While these models show promise in interpreting multimodal data and addressing complex transportation problems, their limitations indicate the need for further refinement in cross-modal alignment, order compositional understanding, fine-detail recognition, object counting accuracy, object hallucination, and computational demands.

This study serves as a roadmap for future research, suggesting taxonomy for the use of MLLMs in end-to-end object detection. It also identifies areas for improvement in model training, computational efficiency, and the broader integration of multimodal data for safer and more efficient transportation systems. By addressing these challenges, MLLMs can play a transformative role in shaping the future of autonomous vehicles and intelligent transportation infrastructure.